\documentclass[10pt,twocolumn,letterpaper]{article}

\usepackage{cvpr}              

\usepackage{comment}
\usepackage{url}
\usepackage{epigraph}
\usepackage{verbatim}
\usepackage{graphicx}
\usepackage{wrapfig}
\usepackage{multirow}
\usepackage{makecell}
\usepackage{color}
\usepackage[table]{xcolor}
\definecolor{mygray}{gray}{0.9}
\usepackage[accsupp]{axessibility}

\definecolor{cvprblue}{rgb}{0.21,0.49,0.74}
\usepackage[pagebackref,breaklinks,colorlinks,citecolor=cvprblue]{hyperref}

\def\paperID{7815} 
\def\confName{CVPR}
\def\confYear{2024}

\title{Learning Multi-dimensional Human Preference for Text-to-Image Generation}

\author{Sixian Zhang\footnotemark[1], Bohan Wang\footnotemark[1], Junqiang Wu\footnotemark[1], Yan Li\footnotemark[2], Tingting Gao, Di Zhang, Zhongyuan Wang\\
Kuaishou Technology}

\begin{document}
\maketitle
\renewcommand{\thefootnote}{\fnsymbol{footnote}}
\footnotetext[1]{These authors contributed equally to this work.}
\footnotetext[2]{Corresponding author.}
\begin{abstract}

Current metrics for text-to-image models typically rely on statistical metrics which inadequately represent the real preference of humans. Although recent work attempts to learn these preferences via human annotated images, they reduce the rich tapestry of human preference to a single overall score. However, the preference results vary when humans evaluate images with different aspects. Therefore, to learn the multi-dimensional human preferences, we propose the Multi-dimensional Preference Score (MPS), the first multi-dimensional preference scoring model for the evaluation of text-to-image models. The MPS introduces the preference condition module upon CLIP model to learn these diverse preferences. It is trained based on our Multi-dimensional Human Preference (MHP) Dataset, which comprises 918,315 human preference choices across four dimensions (i.e., aesthetics, semantic alignment, detail quality and overall assessment) on 607,541 images. The images are generated by a wide range of latest text-to-image models. The MPS outperforms existing scoring methods across 3 datasets in 4 dimensions, enabling it a promising metric for evaluating and improving text-to-image generation. The model and dataset will be made publicly available to facilitate future research. 
Project page: \url{https://wangbohan97.github.io/MPS/}.

\end{abstract}    
\section{Introduction}

\setlength{\epigraphwidth}{0.95\columnwidth}
\renewcommand{\epigraphflush}{center}
\renewcommand{\textflush}{flushepinormal}
\epigraph{\textit{``There are a thousand Hamlets in a thousand people’s eyes."} }
{{\footnotesize{\textit{--Vissarion Belinsky}}}}

Text-to-image generative models~\cite{sd15,sdxl,imagen} have achieved remarkable advancements in recent years, and these models have the capability to generate high-ﬁdelity and contextually relevant images based on textual descriptions (i.e., prompts). 
To evaluate the quality of generated images, several evaluation metrics are proposed, including Inception Score (IS) \cite{IS}, Fréchet Inception Distance (FID) \cite{FID}, and CLIP Score \cite{CLIP_s}. However, these statistical metrics fall short of aligning with human perceptual preferences. For instance, metrics like the IS or the FID, although indicative of image quality to some extent, might not necessarily reflect how a human observer would rate the image in terms of fidelity, coherence, or aesthetic appeal.


\begin{figure}[t]
\begin{centering}
\includegraphics[width=0.975\columnwidth]{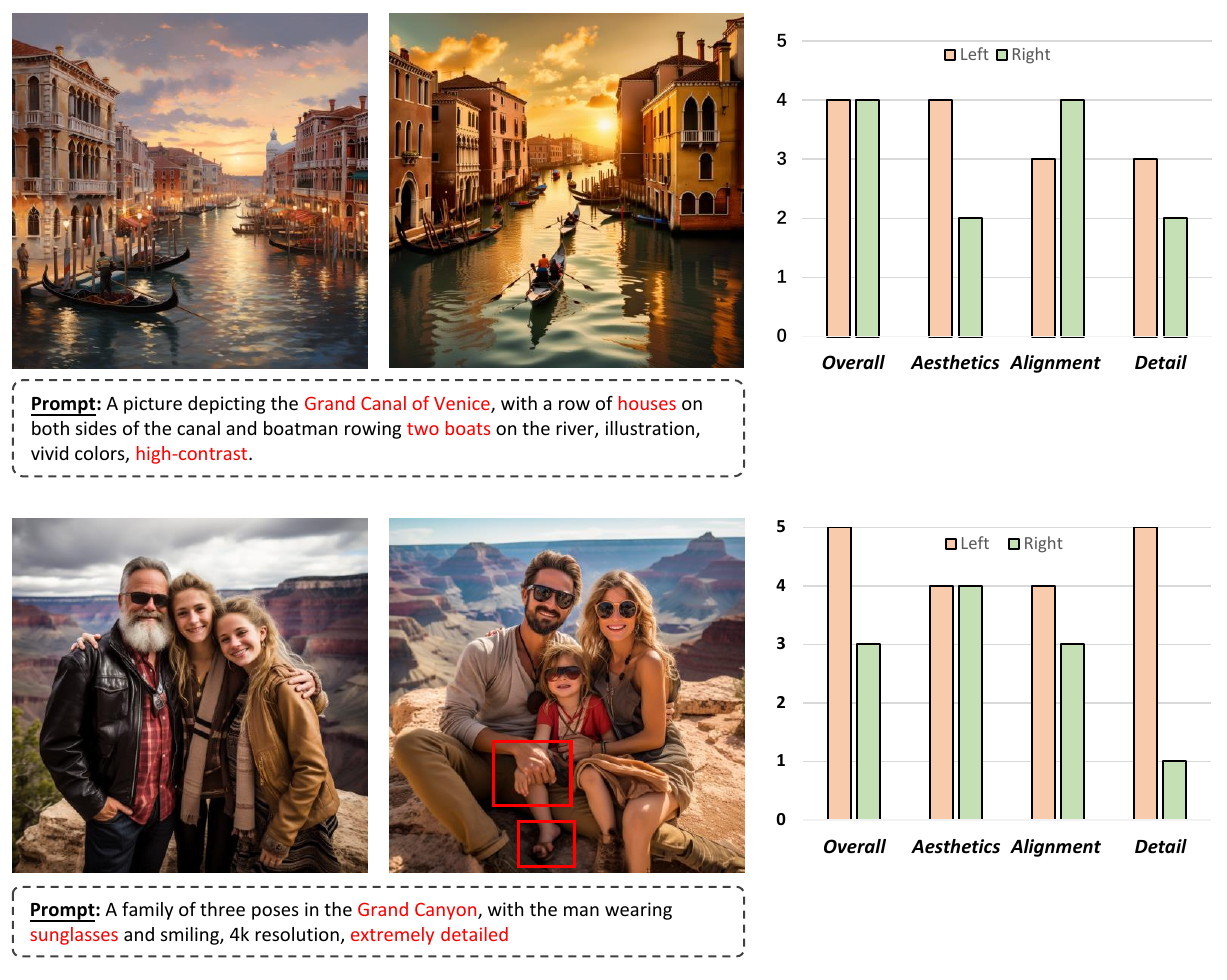}

\end{centering}
\caption{\label{fig:intro}
As humans evaluate images from different perspectives, their preference for the images also varies. Specifically, when examining the images in the top row, the image on the left stands out in terms of aesthetic appeal, though it falls short in semantic alignment (e.g., two boats on the river) compared to its counterpart on the right. In the case of the bottom row, both images are aesthetically pleasing, yet the image on the right is marred by poor detail quality (e.g., as signified by the red bounding boxes around the distorted hand and foot).
}
\vspace{-8pt}
\end{figure}

\begin{table*}[t]
\setlength{\tabcolsep}{2.5pt} \renewcommand{\arraystretch}{1.2}

{\footnotesize{}\caption{\textbf{Comparisons of text-to-image models quality databases}. Our Multi-dimensional Human Preference (MHP) dataset achieves significant advancements over existing work in three aspects, including prompt collection, image generation, and preference annotation. Moreover, it constitutes the largest dataset both in generated images and preference annotations. Note that the Diffusion DB only contains generated images but lacks annotations of human preferences. Besides, KOLORS is an internal dataset derived from in-house platform for designer, which provides $\sim$10w prompts.}
}{\footnotesize\par}
\centering{}{\tiny{}}%
\begin{tabular}{c|cc|cc|cc}
\hline 
\multirow{2}{*}{{\footnotesize{}Dataset}} & \multicolumn{2}{c|}{{\footnotesize{}Prompt collection}} & \multicolumn{2}{c|}{{\footnotesize{}Image Generation}} & \multicolumn{2}{c}{{\footnotesize{}Preference annotation}}\tabularnewline
\cline{2-7} \cline{3-7} \cline{4-7} \cline{5-7} \cline{6-7} \cline{7-7} 
 & {\footnotesize{}Source} & {\footnotesize{}Annotation} & {\footnotesize{}Source} & {\footnotesize{}Number} & {\footnotesize{}Rating} & {\footnotesize{}Dimension}\tabularnewline
\hline 
{\footnotesize{}DiffusionDB \cite{DiffusionDB}} & {\footnotesize{}DiffusionDB} & {\footnotesize{}$\times$} & {\footnotesize{}Diffusion (1)} & {\footnotesize{}1,819,808} & {\footnotesize{}0} & {\footnotesize{}None}\tabularnewline
{\footnotesize{}AGIQA-1K \cite{AGIQA-1K}} & {\footnotesize{}DiffusionDB} & {\footnotesize{}$\times$} & {\footnotesize{}Diffusion (2)} & {\footnotesize{}1,080} & {\footnotesize{}23,760} & {\footnotesize{}Overall}\tabularnewline
{\footnotesize{}PickScore \cite{PickScore}} & {\footnotesize{}Web Application} & {\footnotesize{}$\times$} & {\footnotesize{}Diffusion (3)} & {\footnotesize{}583,747} & {\footnotesize{}583,747} & {\footnotesize{}Overall}\tabularnewline
{\footnotesize{}ImageReward \cite{ImageReward}} & {\footnotesize{}DiffusionDB} & {\footnotesize{}$\times$} & {\footnotesize{}Auto Regressive; Diffusion (6)} & {\footnotesize{}136,892} & {\footnotesize{}410,676} & {\footnotesize{}Overall}\tabularnewline
{\footnotesize{}HPS \cite{HPS}} & {\footnotesize{}DiffusionDB} & {\footnotesize{}$\times$} & {\footnotesize{}Diffusion (1)} & {\footnotesize{}98,807} & {\footnotesize{}98,807} & {\footnotesize{}Overall}\tabularnewline
{\footnotesize{}HPS v2 \cite{HPS_v2}} & {\footnotesize{}DiffusionDB, COCO} & {\footnotesize{}$\checkmark$} & {\footnotesize{}GAN; Auto Regressive; Diffusion, COCO (9)} & {\footnotesize{}430,060} & {\footnotesize{}798,090} & {\footnotesize{}Overall}\tabularnewline
{\footnotesize{}AGIQA-3K \cite{AGIQA-3K}} & {\footnotesize{}DiffusionDB} & {\footnotesize{}$\times$} & {\footnotesize{}GAN; Auto Regressive; Diffusion (6)} & {\footnotesize{}2,982} & {\footnotesize{}125,244} & {\footnotesize{}Overall; Alignment}\tabularnewline
\hline 
{\footnotesize{}MHP} & \thead{\footnotesize{}DiffusionDB, PromptHero,\\ \footnotesize{}KOLORS, GPT4} & {\footnotesize{}$\checkmark$} & {\footnotesize{}GAN; Auto Regressive; Diffusion (9)} & {\footnotesize{}607,541} & {\footnotesize{}918,315} & \thead{\footnotesize{}Aesthetics, Detail,\\ \footnotesize{}Alignment, Overall}\tabularnewline
\hline 
\end{tabular}{\tiny\par}
\vspace{-8pt}
\end{table*}

Contrary to these statistical metrics, several approaches \cite{HPS_v2,PickScore,ImageReward,HPS,AGIQA-1K} turn towards human-centric evaluations, where generated images are manually annotated according to human preferences. Subsequently, models are trained with these annotations to predict preference scores. 
However, these approaches typically utilize a single score to summarize all human preferences, overlooking the multi-dimensionality of human preferences.
As Fig. \ref{fig:intro} shows, the preference results differ when humans evaluate images from various perspectives.
Therefore, using a single-dimensional evaluation method is insufficient in capturing the broad range of personalized needs and preferences. To ensure a comprehensive evaluation of text-to-image synthesis outputs, it is crucial to learn and utilize multi-dimensional human preferences.

To learn the multi-dimensional human preferences, we propose the Multi-dimensional Human Preference (MHP) dataset. Compared to prior efforts \cite{HPS,PickScore,ImageReward,HPS_v2}, the MHP dataset offers significant enhancements in prompts collection, image generation, and preference annotation.
For the prompt collection, previous work \cite{HPS,PickScore,ImageReward} directly utilizes existing open-source datasets (e.g., Diffusion DB \cite{DiffusionDB}) or datasets collated from the internet \cite{PickScore}, overlooking the potential data bias of long-tail distribution. To this end, based on the categories schema of Parti \cite{parti}, we annotate the collected prompts into 7 category labels (e.g., characters, scenes, objects, animals, etc.). For the underrepresented tail categories, we employ Large Language Models (LLMs) (e.g., GPT-4 \cite{GPT-4}) to generate additional prompts. This process results in a balanced prompt collection across various categories, which is used for later image generation.
For image generation, following previous work \cite{HPS_v2,AGIQA-3K}, we not only utilize existing open-source Diffusion models and their variants, but also employ GANs and auto-regressive models to generate images. Consequently, we generate a dataset of 607,541 images, which are further used to create 918,315 pairwise comparisons of images for preference annotation. The quantity of image data constitutes the largest dataset of its kind.
For the annotation of human preferences, contrary to the single annotation of existing work \cite{HPS,HPS_v2,PickScore,AGIQA-1K}, we consider a broader range of dimensions for human preferences and employ human annotators to label each image pair across four dimensions, including aesthetics, detail quality, semantic alignment, and overall score.

To learn human preferences, existing methods employ the pre-trained vision-language models (e.g., CLIP \cite{CLIP}, BLIP \cite{BLIP}) to extract features from images and prompts independently, followed by computing the similarity between them. These methods then fine-tune the network utilizing the collected preference data.
For learning the multi-dimensional preferences, a straightforward strategy is to train separate models for different preferences. 
However, such a simple strategy requires data re-collection and model re-training for the new preference. 
Moreover, due to the potential bias in single-preference data, a model trained under one preference condition often exhibits diminished performance when evaluated against other preferences.
Therefore, we propose the Multi-dimensional Preference Score (MPS), a unified model capable of predicting scores under various preference conditions. Specifically, a certain preference is denoted by a series of descriptive words. For instance, the `aesthetic' condition is decomposed into words such as `light', `color', and `clarity' to describe the attributes of this condition.
These attribute words are used to compute similarities with the prompt, resulting in a similarity matrix that reflects the correspondence between words in the prompt and the specified condition.
On the other hand, features from images and text are extracted using a pre-trained vision-language model. Subsequently, two modalities are fused through a multimodal cross-attention layer. The similarity matrix serves as a mask merged into the cross-attention layer, which ensures that the text only related to the condition is attended to by the visual modality. Then the fused features are used to predict the preference scores.
We evaluate our MPS model on both the existing human preference datasets (i.e., ImageReward \cite{ImageReward} and HPS v2 \cite{HPS_v2}) and our MHP dataset. 
The experimental results indicate that our MPS model surpasses existing benchmarks in evaluating both overall and multi-dimensional preferences, establishing a new state-of-the-art in comparison with related work.

In summary, our main contributions are as follows:
\begin{itemize}
\item We introduce the Multi-dimensional Human Preference (MHP) datasets for evaluating text-to-image models.
The MHP contains balanced prompts and the largest collection of images with multi-dimensional annotations. Based on MHP, we propose a standard test benchmark to evaluate existing text-to-image synthesis models.
\item We propose the MPS model, which learns multi-dimensional human preferences and evaluates the scores of generated images under different preference conditions.
\item Our MPS exhibits superior performance compared to existing methods across three datasets in predicting the overall preferences and multi-dimensional preferences.
\end{itemize}
\section{Related work}
\subsection{Text-to-image Generation and Evaluation}



The text-to-image task aims to synthesize realistic images from natural language description (i.e., prompt). Several work attempts to tackle this problem, including GANs~\cite{lafite, vqgan}, auto-regressive~\cite{cogview,cogview2,parti} and diffusion models~\cite{sd15,sdxl,imagen}. 
Among the previously mentioned methods, diffusion models gain significant attention for their exceptional performance. These methods are principally divided into two categories: latent-based and pixel-based approaches. The Latent Diffusion Model (LDM)~\cite{sd15} is notable as the first to introduce a latent-based diffusion model, leveraging an auto-encoder to map images into a latent space where the diffusion process is executed. Following this, Stable Diffusion has notably propelled the field forward by open-sourcing SD series~\cite{sdxl}. In contrast, DALL·E 2~\cite{dalle2} and Imagen~\cite{imagen} are predicated on pixel-based diffusion models. Besides, the Imagen~\cite{imagen} integrates the large language model T5 XXL to achieve a text-to-image super-resolution diffusion model capable of producing highly realistic images.
Current text-to-image models excel in creating high-quality images but often miss aligning with human preferences in real-world applications.
For evaluating text-to-image models, several evaluation metrics are proposed to evaluate the quality of generated images, including Inception Score (IS)~\cite{IS}, Fréchet Inception Distance (FID)~\cite{FID}, and CLIP Score~\cite{CLIP_s}. However, these statistical metrics fall short of aligning with human perceptual preferences. Our MPS provides a comprehensive evaluation for text-to-image generations, facilitating the evaluation of alignment with multi-dimensional human preference.



\subsection{Learning human preferences}

Currently, several studies attempt to collect and learn human preferences for the evaluation of text-to-image generation. They use the collected data to fine-tune visual-language models (VLMs) to align with human selections.

HPS \cite{HPS} introduces the HPD dataset of human preference choices. The images are generated solely using the Stable Diffusion model with prompts from Diffusion DB. They train HPS model utilizing the human preference annotations of HPD to align it with human preference. Subsequently, they fine-tune the Stable Diffusion model under the guidance of HPS, leading to better generated images that are more preferred by human users. However, HPS is limited to a single generation model and a relatively small number of images. Furthermore, HPS v2 \cite{HPS_v2} introduces a larger dataset, employing 8 generative models, including Diffusion models, GANs, and Auto regressive models, and also incorporating captions from the COCO dataset. 
However, both HPS and HPS v2 primarily focus on overall human preferences, not considering the diversity of human tastes.

\begin{figure}[t]
\begin{centering}
\includegraphics[width=1.02\columnwidth]{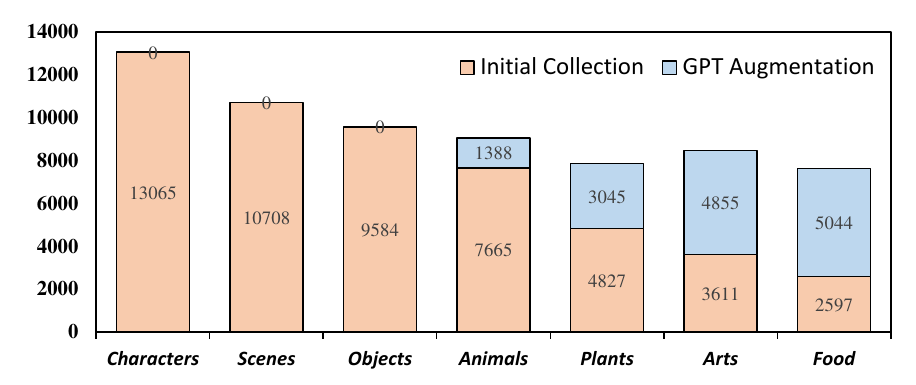}

\end{centering}
\vspace{-4pt}
\caption{\label{fig:database}
\textbf{Prompt collection}. The initially collected prompts exhibit a long-tail distribution across various categories. After prompt augmentation with GPT, we obtain a relatively balanced prompt dataset, which contains 66,389 prompts.
}
\vspace{-8pt}
\end{figure}

PickScore \cite{PickScore} proposes a web application designed to collect prompts and human preference annotations from real users. 
Unlike previous methods adopting prompts from existing datasets (e.g. DiffusionDB), the prompts of PickScore are directly generated by actual users. The dataset of PickScore is sizeable, however, it focuses only on the overall preferences, lacking detailed annotations for multi-dimension preferences.

ImageReward \cite{ImageReward} employs four types of Diffusion models along with an auto-regression-based model. Their annotation of generated images is more detailed with scoring ranging from 0 to 7. Beyond overall satisfaction, they also consider annotations for alignment and fidelity. 
However, they merge alignment and fidelity into a single overall score, inadequately capturing the multi-dimensional dimensions of human preferences.

AGIQA-1k \cite{AGIQA-1K} and AGIQA-3k \cite{AGIQA-3K} utilize a range of generative models, including Diffusion, GAN, and Auto regressive models, to produce images. They consider both overall and alignment preferences. However, their dataset size is considerably smaller compared to existing work.

Our MHP dataset represents an advancement over previous work in terms of prompt collection, image generation, and preference annotation. 
\section{MHP Dataset}
\subsection{Prompt collection and annotation}

\begin{table}[t]
\setlength{\tabcolsep}{2.5pt} \renewcommand{\arraystretch}{1}

\caption{\label{tab:Image-sources}\textbf{Image generation}. The image sources of the MHP dataset consist of the images generated from 9 text-to-image generative models. 
Note that KOLORS is an internal model for in-house designer platform.}

\centering{}%
\begin{tabular}{c|ccc}
\hline 
{\small{}Source} & {\small{}Type} & {\small{}Images} & {\small{}Split}\tabularnewline
\hline 
{\small{}KOLORS} & {\small{}Diffusion} & {\small{}211,707} & {\small{}Train and test}\tabularnewline
{\small{}DeepFloyd IF} & {\small{}Diffusion} & {\small{}27,311} & {\small{}Train and test}\tabularnewline
{\small{}Stable Diffusion XL} & {\small{}Diffusion} & {\small{}89,176} & {\small{}Train and test}\tabularnewline
{\small{}Openjourney v4} & {\small{}Diffusion} & {\small{}133,875} & {\small{}Train and test}\tabularnewline
{\small{}Stable Diffusion v2.0} & {\small{}Diffusion} & {\small{}84,590} & {\small{}Train and test}\tabularnewline
{\small{}Stable Diffusion v1.5} & {\small{}Diffusion} & {\small{}56,882} & {\small{}Train and test}\tabularnewline
{\small{}VQGAN+CLIP} & {\small{}GAN} & {\small{}1,000} & {\small{}Test}\tabularnewline
{\small{}LAFITE} & {\small{}GAN} & {\small{}1,000} & {\small{}Test}\tabularnewline
{\small{}CogView2} & {\small{}Autoregressive} & {\small{}2,000} & {\small{}Test}\tabularnewline
\hline 
\end{tabular}
\vspace{-8pt}
\end{table}

Our prompts are carefully collected from several databases, including PromptHero\cite{PromptHero}, DiffusionDB\cite{DiffusionDB} and KOLORS-dataset (an internal dataset derived from in-house platform for designers).
Following the category schema from Parti\cite{parti}, we further merge some categories, and finally determine 7 categories as illustrated in Fig. \ref{fig:database}. The definitions and merging rules are detailed in the supplementary material.
Based on these defined categories, we employ human annotators to label the initially collected 59,396 prompts. Additionally, the annotators are also required to filter out anomalous prompts, such as those that are incoherent, incomprehensible, or have punctuation errors. After annotation, we obtain 52,057 prompts. The distribution of these prompts, based on their categories, is depicted in Fig. \ref{fig:database}. As the figure shows, the initially collected prompts exhibit a long-tail distribution across categories. Such category imbalances might lead to imbalanced generated images. Consequently, the human preferences learned from these imbalanced data could also be biased. As a result, we further expand our prompts.

By employing the GPT-4\cite{GPT-4}, we obtain additional prompts to supplement categories with initially low quantities (see supplementary materials for more examples of the generated prompts). These generated prompts are further refined by annotators to remove those incoherent or incomprehensible items. 
As shown in Figure \ref{fig:database}, after supplementation, we obtain 66,389 prompts and the distribution of prompts across categories is balanced. These balanced prompts help us in learning more representative human preferences.

\begin{figure}[t]
\begin{centering}
\includegraphics[width=0.93\columnwidth]{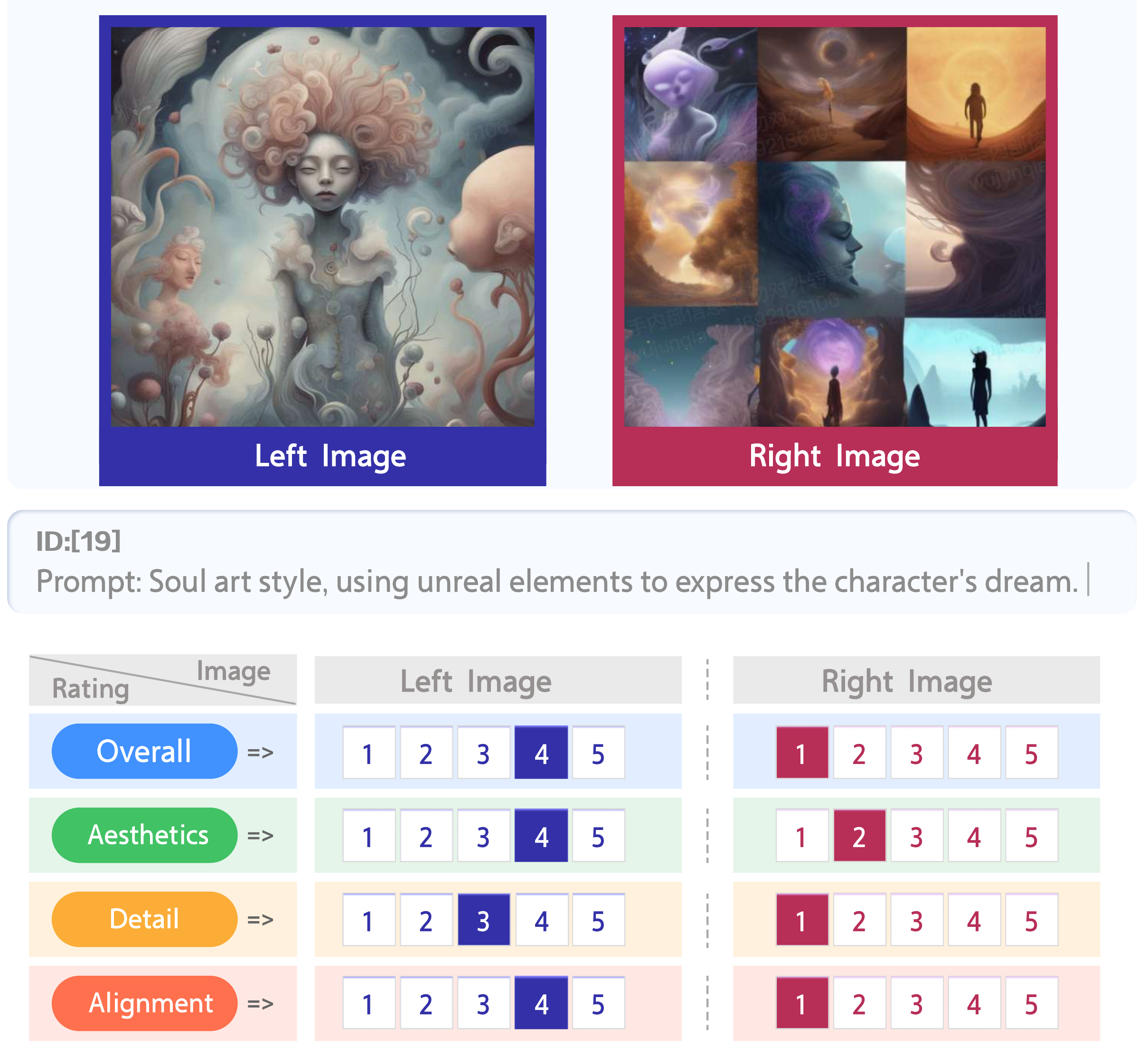}

\end{centering}
\caption{\label{fig:interface}
\textbf{Annotation interface}. 
Annotators are required to evaluate the preference for the given image pair on four dimensions, including aesthetics, detail quality (detail), semantic alignment (alignment) and overall score (overall)).
Annotation scores are discrete values ranging from 1 to 5, which are subsequently normalized to Boolean values of 0 or 1. When the scores are tied, the normalized score is set to 0.5.
}
\vspace{-8pt}
\end{figure}

\subsection{Image collection and annotation}

As shown in Table~\ref{tab:Image-sources}, we utilize Diffusion models (such as the Stable Diffusion series and DeepFloyd IF), GANs, and AutoRegressive models to generate images based on the obtained prompts. Each model produces 2-4 images for every single prompt. The generated images come from a variety of model architectures, with various image resolution scales (e.g., $512 \times 512$, $1024 \times 1024$, $1366 \times 768$), and aspect ratios (e.g., 1:1, 16:9). This diversity ensures a comprehensive representation of the text-to-image models' generalization capability.

Images generated from the same prompt are paired together for comparison. To enhance the representativeness of these image pairs, the construction of image pairs sources not only from images generated by different models but also includes those produced by the same model using different random seeds. Based on these contrastive image pairs, we employ human annotators to evaluate the image pairs with our annotation interface as shown in Fig. \ref{fig:interface}. The annotators are required to evaluate the quality of the generated image pairs based on three sub-dimensions (i.e., aesthetics, text-image consistency, and detail) and one overall dimension (i.e., overall score). These four dimensions are defined as follows:

\begin{enumerate}
    \item \textbf{Aesthetics}: annotators should measure the aesthetic quality of a generated image pair in terms of composition, light contrast, color matching, clarity, tone, style, depth of field, atmosphere, and artistry of the image.

    \item \textbf{Detail quality}: annotators should focus on the delicacy of image details such as texture, hair, and light and shadow, whether there is distortion in the face, hands, and limbs of the characters, whether there is a blurry overall view, object distortion, severe deformation.

    \item \textbf{Semantic alignment}: annotators should evaluate the semantic consistency of the generated images with the prompts, and the evaluation includes measuring whether the generated image accurately matches the textual description (e.g., quantity, attributes, location, positional relationships) and whether there is missing or redundant content in the generated image. 

    \item \textbf{Overall assessment}: Based on the combination of above aspects and subjective preferences, the annotators assess the quality of each generated image from a holistic perspective.
\end{enumerate}
The annotators rate all these scores of each image in the pair into five distinct levels (from 1 to 5), and the scores are eventually normalized to $[0,1]$.
The image annotation is completed by a crowdsourced team of 210 members. 
Before the official annotation, each member needs to perform pre-annotation, where any member whose annotation results have a high degree of inconsistency with that of the majority is disqualified. 
Eventually, 198 members participate in the annotation of generated image pairs, of which 170 members act as annotators and 28 members act as quality inspectors. 
Each image pair is annotated by three annotators respectively and the final result is averaged from these three annotation results.  
20\% of the annotated data is extracted and sent to the quality inspector for inspection.
If there is significant difference in annotation results between annotators and inspectors, the annotated data is considered invalid and will be relabeled.

\subsection{Statistics}

In summary, we collect 66,389 prompts and employ 9 recent text-to-image models to generate 607,541 images. Based on these images, we construct 918,315 image pairs. Notably, 20\% of these pairs are created using the same model but with different settings, while the other 80\% are produced by different models, allowing for a wide range of comparisons. Each image pair is then annotated by 3 distinct annotators across 4 dimensions to enable the study of diversity in human preference.

We divide the annotated data into training, validation and test sets. The training and validation set contains 898,315 and 10,000 image pairs and the test set comprises 10,000 pairs. To ensure that the data distribution is representative, the test set includes not only images generated by Diffusion models but also those produced by GAN and Autoregressive models.

\begin{figure}[t]
\begin{centering}
\includegraphics[width=0.93\columnwidth]{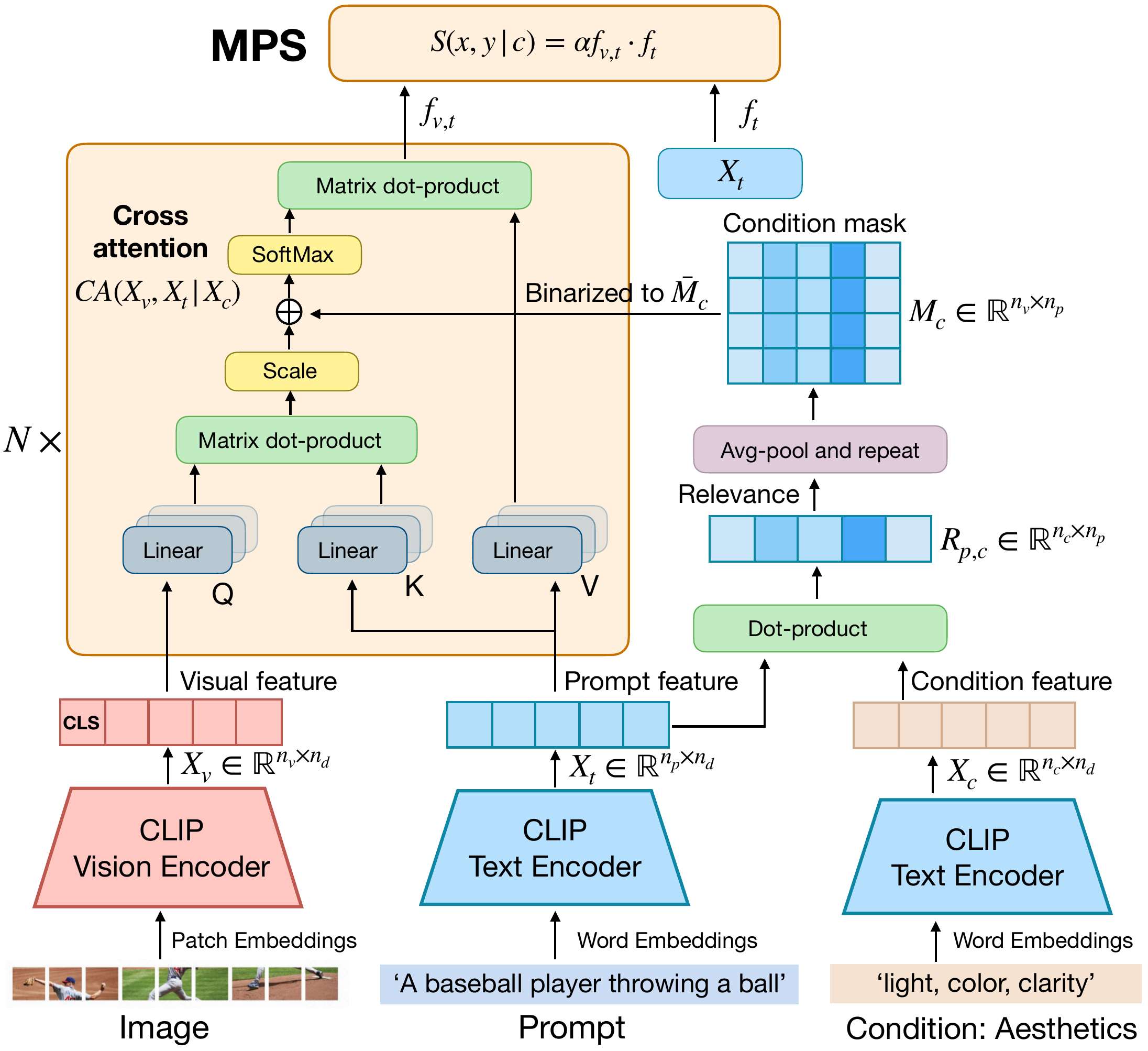}

\end{centering}
\caption{\label{fig:framework}
\textbf{The framework of Multi-dimensional Preference Score (MPS)}.
The MPS takes the generated image, prompt and preference condition as the input, and predicts the quality (i.e. human preference) of the generated image under the given preference condition.
}
\vspace{-8pt}
\end{figure}

\section{Multi-dimensional Preference Prediction}
\subsection{Model Structure of MPS}
As shown in Fig. \ref{fig:framework}, we adopt CLIP \cite{CLIP} to initially extract features from images and prompts. Given a prompt $x$, a generated image $y$ and a preference condition $c$, the visual feature of $y$ is obtained by the vision encoder of the CLIP by $X_v=E_{v}(y)$, where $X_v \in \mathbb{R}^{n_v \times n_d}$. $n_v$ is the token number of the image and $n_d$ is the feature dimension. $X_t=E_{t}(x)$, where $X_t \in \mathbb{R}^{n_p \times n_d}$ and $n_p$ is the token number of the text. 
$X_c=E_{c}(c)$, where $X_c \in \mathbb{R}^{n_c \times n_d}$ and $n_p$ is the token number of word sets representing the preference. For the setup of word sets for each preference, please refer to Sec. \ref{sec:setup}.
Previous works \cite{PickScore,HPS,ImageReward} typically utilize the first dimension of $X_v$ and the last dimension of $X_t$ to calculate the preference score, which loses a lot of detailed information of both image and text. Alternatively, we employ the full range of $X_v$ and $X_t$ and fuse two modalities through the Cross Attention (CA) module 
\begin{equation}
CA\left(X_{v},X_{t}\right)=\sigma\left(\frac{X_{v}W_{q}\left(X_{t}W_{k}\right)^{T}}{\sqrt{n_d}}\right)X_{t}W_{v}
\label{eq:cross}
\end{equation}
where $\sigma$ is the SoftMax activation function, and $W_* \in \mathbb{R}^{n_d \times n_d}$ are parameters and biases are omitted. 
Our motivation is computing the preference scores in different preference conditions. 
Specific words in the prompt should be given more attention based on different conditions, e.g., when aesthetics is considered as a condition, words in the prompt related to color, light, and clarity should be taken into account more when calculating the score. Therefore, we propose a condition mask to highlight the relevant tokens while suppressing the irrelevant tokens. 
The condition is represented by a series of attribute words, e.g. the preference condition of `Aesthetic' is represented by a set of words, including light, color and clarity. 
The relevance of prompt and condition is computed by $R_{p,c}=X_{c}X_{t}^{T}W_c+b_c$, where $R_{p,c} \in \mathbb{R}^{n_c\times n_p}$ and $W_c$ and $b_c$ are learnable parameters. The $R_{p,c}$ is averaged along the dimension $n_c$ and then repeated $n_v$ times to obtain the mask $M_c\in\mathbb{R}^{n_v\times n_p}$. 
The $M_c$ is further binarized to $\bar{M}_c$, where elements below the similarity threshold are assigned to negative infinity and others are set to zero.
Based on the condition mask $\bar{M}_c$, the Eq. \ref{eq:cross} is further improved as
\begin{equation}
CA\left(X_{v},X_{t}|X_c\right)=\sigma\left(\frac{X_{v}W_{q}\left(X_{t}W_{k}\right)^{T}}{\sqrt{d}}+\bar{M}_c\right)X_{t}W_{v}
\label{eq:cross condition}
\end{equation}
The Eq. \ref{eq:cross condition} ensures that the parts of the prompt that are relevant to the condition information receive more attention when computing the preference score.
We adopt the first dimension (i.e. the cls token) of fused feature $CA\left(X_{v},X_{t}|X_c\right)$ for further prediction, which is denoted as $f_{v,t}$ and $s\in\mathbb{R}^{1\times n_d}$
Additionally, to prevent the issue of excessively short prompts leading to a scenario where no words of the prompt are related to the condition.
In such a scenario, the cross-attention layer would be unable to capture the information from the prompt. 
Therefore, we supplement the $f_{v,t}$ with additional prompt feature $f_t$, where $f_t$ is the last dimension of $X_t$. 
Consequently, the MPS is obtained by
\begin{equation}
S(x,y|c)=\alpha f_{v,t} \cdot f_t
\label{eq:MPS}
\end{equation}
where $\alpha$ is a learned scalar while $x$, $y$ and $c$ denote the prompt, image and preference condition, respectively.

\begin{table}
\setlength{\tabcolsep}{2.5pt} \renewcommand{\arraystretch}{1.2}

\caption{
Main results of MPS and comparison methods on human preference evaluation. Preference accuracy (\%) is calculated on ImageReward, HPD v2, and our MHP dataset. 
}
\label{tab:Comparison on overall}
\centering{}{\small{}}%
\begin{tabular}{c|c|ccc}
\hline 
{\small{}ID} & {\small{}Preference Model} & {\small{}ImageReward} & {\small{}HPD v2} & {\small{}MHP (Overall)}\tabularnewline
\hline 
{\small{}1} & {\small{}CLIP score \cite{CLIP}} & {\small{}54.3} & {\small{}71.2} & {\small{}63.7}\tabularnewline
{\small{}2} & {\small{}Aesthetic Score \cite{LAION-5B}} & {\small{}57.4} & {\small{}72.6} & {\small{}62.9}\tabularnewline
{\small{}3} & {\small{}ImageReward \cite{ImageReward}} & {\small{}65.1} & {\small{}70.6} & {\small{}67.5}\tabularnewline
{\small{}4} & {\small{}HPS \cite{HPS}} & {\small{}61.2} & {\small{}73.1} & {\small{}65.5}\tabularnewline
{\small{}5} & {\small{}PickScore \cite{PickScore}} & {\small{}62.9} & {\small{}79.8} & {\small{}69.5}\tabularnewline
{\small{}6} & {\small{}HPS v2 \cite{HPS_v2}} & {\small{}65.7} & {\small{}83.3} & {\small{}65.5}\tabularnewline
\hline 
{\small{}7} & {\small{}MPS (Ours)} & \textbf{\small{}67.5} & \textbf{\small{}83.5} & \textbf{\small{}74.2}\tabularnewline
\hline 
\end{tabular}{\small\par}
\vspace{-8pt}
\end{table}

\subsection{Training}
The input for our objective includes our scoring function MPS $S(x,y|c)$, a prompt $x$, two generated image $y_1$, $y_2$, a preference condition $c$ and the preference score (annotated by human) $p$, where $p$ takes a value of $[1,0]$ for $y_1$ is preferred, $[0,1]$ if $y_2$ is preferred, or $[0.5, 0.5]$ for ties. 
Following previous work \cite{PickScore}, the training objective minimizes the KL-divergence between the annotation $p$ and the softmax-normalized prediction
\begin{equation}
\begin{split}
&\hat{p}_{i,c}=\frac{expS\left(x,y_{i}|c\right)}{\sum_{i=1}^{2}expS\left(x,y_{i}|c\right)}\\
&L_{P} = \sum_{c}\sum_{i=1}^{2}p_{i,c}(log p_{i,c} - log \hat{p}_{i,c})
\end{split}
\end{equation}
We initialize the text and vision encoders, \( E_t \) and \( E_v \), with parameters from the pre-trained CLIP-H model, while the remaining parameters are subject to random initialization.
We train our MPS on MHP datasets for 30,000 steps, with a batch size of 128, a learning rate of 3e-6, and a warmup period of 500 steps.

\section{Experiments}

\subsection{Experimental Setup \label{sec:setup}}
\paragraph{Preference condition setting.}
We utilize the following collection of word sets to represent human preferences:
1) Aesthetics: \textit{light, color, clarity, tone, style, ambiance, artistry};
2) Detail quality: \textit{shape, face, hair, hands, limbs, structure, instance, texture};
3) Semantic alignment: \textit{quantity, attributes, position, number, location};
4) Overall: \textit{light, color, clarity, tone, style, ambiance, artistry, shape, face, hair, hands, limbs, structure, instance, texture, quantity, attributes, position, number, location}.
\paragraph{Evaluation setting.}
We select widely used statistical metrics for evaluating text-to-image models, namely the CLIP score \cite{CLIP_s} and Aesthetic score \cite{LAION-5B} for comparison. Additionally, we also choose methods that align with human preferences for evaluating text-to-image models, including Image Reward \cite{ImageReward}, HPS \cite{HPS,HPS_v2} and PickScore \cite{PickScore}.
Following previous works \cite{HPS_v2,PickScore}, we utilize publicly available pre-trained models without finetuning for evaluation.

\begin{table}
\setlength{\tabcolsep}{2.9pt} \renewcommand{\arraystretch}{1.2}

\caption{The evaluation of MPS and related scoring functions for the prediction of multi-dimensional human preferences(\%).}
\label{tab:multinational preference}
\centering{}%
\begin{tabular}{c|c|cccc}
\hline 
{\small{}ID} & {\small{}Preference Model} & {\small{}Overall} & {\small{}Aesthetics} & {\small{}Alignment} & {\small{}Detail}\tabularnewline
\hline 
{\small{}1} & {\small{}CLIP score \cite{CLIP}} & {\small{}63.67} & {\small{}68.14} & {\small{}82.69} & {\small{}61.71}\tabularnewline
{\small{}2} & {\small{}Aesthetic Score \cite{LAION-5B}} & {\small{}62.85} & {\small{}82.85} & {\small{}69.36} & {\small{}60.34}\tabularnewline
{\small{}3} & {\small{}ImageReward \cite{ImageReward}} & {\small{}67.45} & {\small{}74.79} & {\small{}75.27} & {\small{}58.31}\tabularnewline
{\small{}4} & {\small{}HPS \cite{HPS}} & {\small{}65.51} & {\small{}73.86} & {\small{}73.86} & {\small{}62.05}\tabularnewline
{\small{}5} & {\small{}PickScore \cite{PickScore}} & {\small{}69.52} & {\small{}70.95} & {\small{}70.92} & {\small{}56.74}\tabularnewline
{\small{}6} & {\small{}HPS v2 \cite{HPS_v2}} & {\small{}65.51} & {\small{}73.86} & {\small{}73.87} & {\small{}62.06}\tabularnewline
\hline 
{\small{}7} & {\small{}MPS (Ours)} & \textbf{\small{}74.24} & \textbf{\small{}83.86} & \textbf{\small{}83.87} & \textbf{\small{}85.18}\tabularnewline
\hline 
\end{tabular}
\vspace{-8pt}
\end{table}

\subsection{Evaluation Results}
\paragraph{Overall Preference accuracy.}
Previous works on learning human preferences mostly focus on a singular, overall preference, i.e., summarizing human preferences with an overall score. For a fair comparison, we choose existing publicly available human preference datasets: the ImageReward test set \cite{ImageReward} and the HPD v2 test set \cite{HPS_v2}, along with our MHP dataset (using only data annotated with overall scores) to compare our method with relevant baselines. As shown in Tab. \ref{tab:Comparison on overall}, our MPS demonstrates a better accuracy across these three datasets, indicating the strong generalization capability of our method.

\begin{figure}[t]
\begin{centering}
\includegraphics[width=1.02\columnwidth]{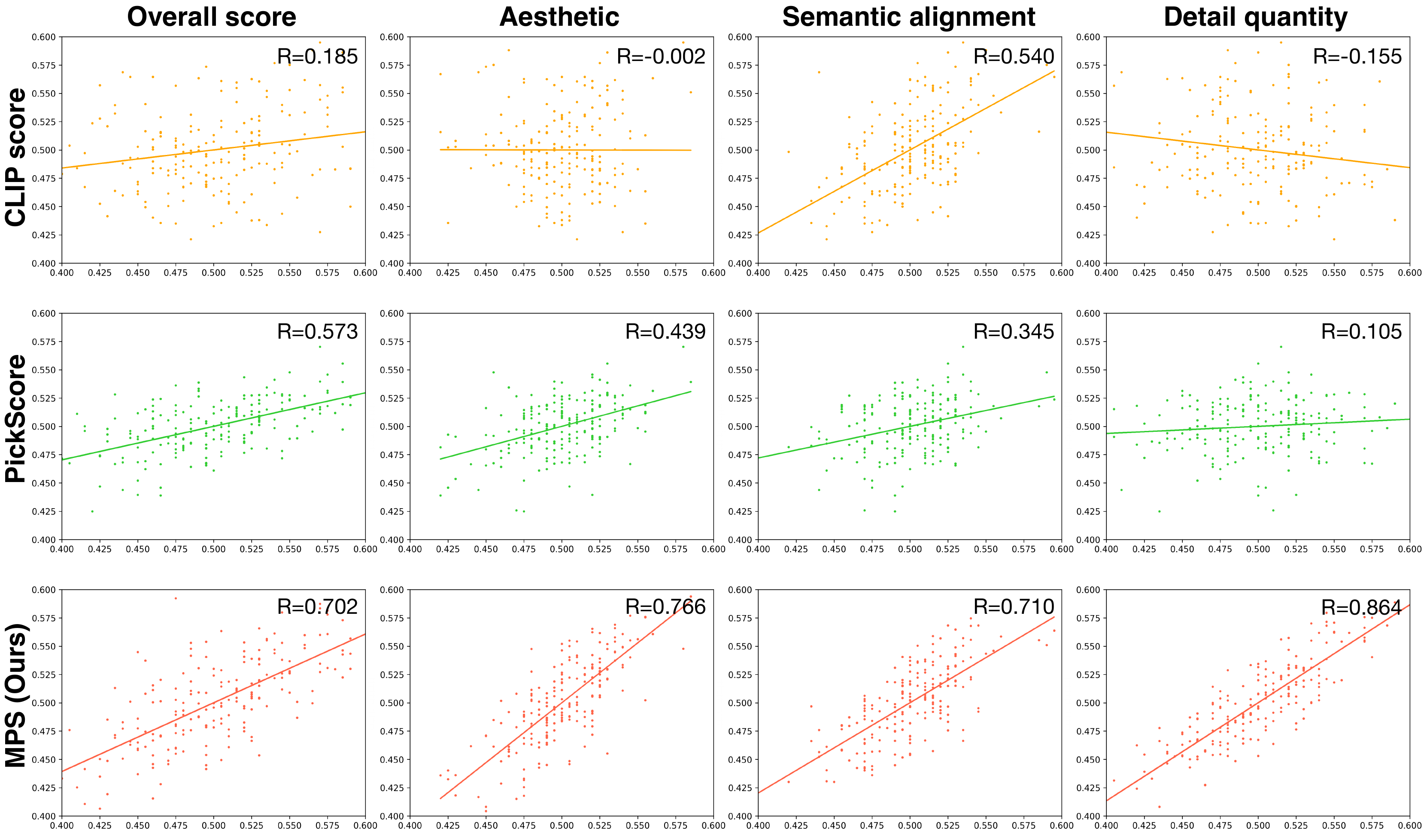}
\par\end{centering}
\caption{\textbf{Correlation between real user preferences and model predictions}. The x-axis of each subplot represents the annotated real human preferences, and the y-axis denotes the model's predictions. We examine three models: CLIP score, PickScore, and MPS (ours). Each subplot is annotated with the calculated correlation coefficient R-value, where a higher R-value indicates a closer alignment of the model's predictions with actual human preferences.
}
\label{fig:correlation experiments}
\vspace{-8pt}
\end{figure}

\paragraph{Multinational Preference accuracy.}
In addition to the overall score, we also compare the performance of previous works and our method in predicting multi-dimensional human preferences based on our MHP dataset. As Tab. \ref{tab:multinational preference} illustrates, the CLIP Score and Aesthetic score focus on specific types of preferences, only perform well in certain preferences (e.g., semantic alignment or aesthetics). However, they fall short in predicting other preferences compared to models trained on human preferences. 
Additionally, preference models \cite{HPS,HPS_v2,ImageReward,PickScore} generally perform better in overall score and some other dimensions but lack generalization in certain specific preferences (e.g., details)\footnote{Since these methods have been trained solely to generate an overall score, we could only duplicate it for multinational preferences.}.
We illustrate the Fig. \ref{fig:correlation experiments} to reveal the underlying reasons for this poor generalization.
In the first and second rows of Fig. \ref{fig:correlation experiments}, 
the score functions exhibit a high correlation only with the trained preferences (e.g. semantic alignment and overall score), but perform poorly on other preferences. It is important to note that the prediction of different preferences is based on the same data, albeit with different preference annotations. This indicates that not all preferences are strongly correlated, which results in that improvements in one preference might come at the expense of others.
Therefore, only learning a single score is inadequate in fully reflecting the complexity of human preferences.
In contrast, our MPS learns human preferences with the condition mask from multiple dimensions and maintains high consistency across all dimensions of human preferences, as shown in the third row of Fig. \ref{fig:correlation experiments}. Besides, as Tab. \ref{tab:multinational preference} indicated, MPS outperforms the related works by a large margin in predicting the multi-dimensional human preference on the MHP dataset.


\begin{figure*}[t]
\begin{centering}
\includegraphics[width=0.99\textwidth]{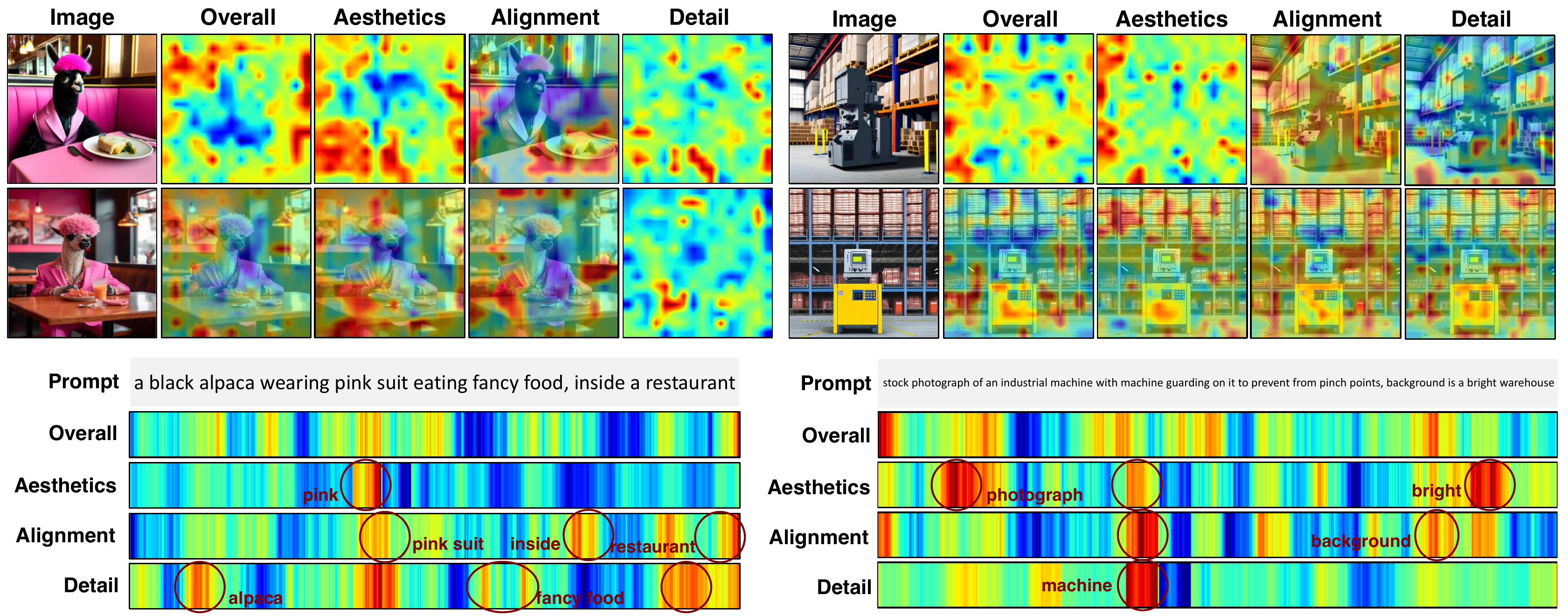}
\par\end{centering}
\caption{\textbf{Visualization}. We leverage Grad-CAM and the condition mask \( M_c \) to visualize attention heatmaps for the image and prompt. The condition mask results in the prompt focusing on words related to the preference condition. For Aesthetic, the model tends to focus on colors (e.g., \textit{pink}) and lighting (e.g., \textit{bright}). In the case of semantic alignment, the focus shifts to attributes (e.g., \textit{pink suit}) and position (e.g., \textit{inside}, \textit{background}). For detail quantity, the model's attention is on instances (e.g., \textit{alpaca}, \textit{machine}, \textit{fancy food}). On the image side, the model also tends to focus on areas of the image that correlate with the parts of the prompt receiving attention.
}
\label{fig:visualization}
\vspace{-8pt}
\end{figure*}

\paragraph{Visualization Results.}
Further, we aim to explore why our MPS exhibits strong generalization across various dimensions of human preferences, even some preferences (e.g., detail quantity) are not highly correlated with others. To this end, we visualized the attention map of images and prompts that MPS focuses on when predicting human preferences. 
As shown in Fig. \ref{fig:visualization}, we employ Grad-CAM \cite{Grad-CAM} and \( f_{v,t} \) to generate attention heatmaps of the image, and utilize the values of \( M_c \) to represent the attention heatmap of the prompt.
The visualization results indicate that our HPS attends to different regions of prompts and images depending on the specific preference condition. 
This is attributed to the condition mask, which allows only those words in the prompt related to the preference condition to be observed by the image. The condition mask ensures that the model predicts the preference with different inputs, and the model only needs to calculate the similarity between patches in the image and the retained partial prompt to determine the final score.
Therefore, the selective focus enabled by the condition mask allows utilizing a unified model to predict multinational preferences effectively, even if some preferences have weak correlations with others.

\paragraph{Ablation study.}
We conduct ablation studies to verify the effectiveness of each component, as illustrated in Tab. \ref{tab:ablation}. Compared to the baseline\footnote{Note that the baseline shares the same network architecture with PickScore in Tab. \ref{tab:Comparison on overall} and \ref{tab:multinational preference}, but is trained on our MHP dataset.}  (i.e., PickScore), the cross-attention module enables more comprehensive integration between image and prompt, leading to improvement of prediction accuracy in overall score, aesthetics, and semantic alignment. 
However, the model still underperforms in detail quantity, which is less correlated with other preferences. 
The addition of the condition mask $M_c$ alleviates this issue and improves the prediction performance across various preferences, especially in the detail quantity.
Furthermore, we train separate models for different preferences. 
Experimental results indicate that models trained separately for each preference do not perform as well as those trained to learn multiple preferences simultaneously. 
We infer that more extensive annotations and unified training contribute to better model generalization. Ablation studies validate the effectiveness of each module, particularly the condition mask, in learning multiple preferences.


\begin{table}
\setlength{\tabcolsep}{2pt} \renewcommand{\arraystretch}{1.2}\label{ablations}

\caption{
\textbf{Ablation study for different modules used in MPS}. Base: PickScore is employed as the baseline model. CA: Cross-attention module. Mask: Preference condition mask
$M_{c}$. Row 4 illustrates the models that have identical structures but are trained separately for each preference type. 
}
\label{tab:ablation}

\centering{}%
\begin{tabular}{c|ccc|cccc}
\hline 
\multirow{2}{*}{{\small{}ID}} & \multicolumn{3}{c|}{{\small{}Module}} & \multirow{2}{*}{{\small{}Overall}} & \multirow{2}{*}{{\small{}Aesthetics}} & \multirow{2}{*}{{\small{}Alignment}} & \multirow{2}{*}{{\small{}Detail}}\tabularnewline
\cline{2-4} \cline{3-4} \cline{4-4} 
 & {\small{}Base} & {\small{}CA} & {\small{}Mask} &  &  &  & \tabularnewline
\hline 
{\small{}1} & {\small{}$\checkmark$} &  &  & {\small{}70.65} & {\small{}71.05} & {\small{}70.81} & {\small{}57.46}\tabularnewline
{\small{}2} & {\small{}$\checkmark$} & {\small{}$\checkmark$} &  & 74.11 & 71.83 & 73.02 & 58.79\tabularnewline
{\small{}3} & {\small{}$\checkmark$} & {\small{}$\checkmark$} & {\small{}$\checkmark$} & {\small{}\textbf{74.24}} & {\small{}\textbf{83.86}} & {\small{}\textbf{83.87}} & {\small{}\textbf{85.18}}\tabularnewline
\hline 
{\small{}4} & \multicolumn{3}{c|}{{\footnotesize{}\cellcolor{mygray}Separately trained MPS}} & {\small{}\cellcolor{mygray}73.51} & {\small{}\cellcolor{mygray}80.54} & {\small{}\cellcolor{mygray}79.68} & {\small{}\cellcolor{mygray}76.81}\tabularnewline
\hline 
\end{tabular}
\vspace{-8pt}
\end{table}

\subsection{MPS Benchmark}
Based on the MPS model and the collected MHP dataset, we introduce the MPS benchmark for evaluating text-to-image models across multiple dimensions. The MPS benchmark includes a set of evaluation prompts designed to assess the models on a total of 4,000 prompts, covering seven categories: characters, scenes, objects, animals, plants, arts, and food.
Each category comprises 500 prompts. Our MPS assesses the images generated by the text-to-image models across four dimensions: Aesthetic, Semantic Alignment, Detail Quantity, and Overall Score. The scoring results can assist users in selecting superior models based on their personal preferences. Additionally, the scoring results can also enhance the generative models' performance by selecting more preferable images with higher MPS scores.


\section{Conclusions}

In this work, we introduce the Multi-dimensional Human Preference (MHP) dataset and Multi-dimensional Preference Score (MPS) to evaluate text-to-image models from multi-dimensional human preferences.
The MHP dataset offers improvements over previous methods in prompt collection, image generation, and preference annotation, which comprises 918,315 human preference choices across four dimensions. These preferences include aesthetics, semantic alignment, detail quantity, and overall score. 
Additionally, to align the multi-dimensional human preferences, we propose the MPS, which employs a unified network to score generated images based on varying preference conditions. 
The MPS introduces a condition mask that retains words in the prompt related to the preference condition. Subsequently, the model integrates only the retained prompt with the image to compute the final score. 
MPS outperforms related works in predicting multi-dimensional human preferences across three datasets, which demonstrates the generalization of our method.

\paragraph*{{\small{}Acknowledgements:}}
\noindent {\small{}
We sincerely thank Zhuang Li, Lingyu Zou, and Peihan Li for their valuable discussions and feedback.
}{\small\par}

{
    \small
    \bibliographystyle{ieeenat_fullname}
    \bibliography{main}
}


\end{document}


\clearpage
\setcounter{page}{1}
\maketitlesupplementary

\section{Text-to-Image Ranking}

\begin{figure}[t]
\begin{centering}
\includegraphics[width=0.96\columnwidth]{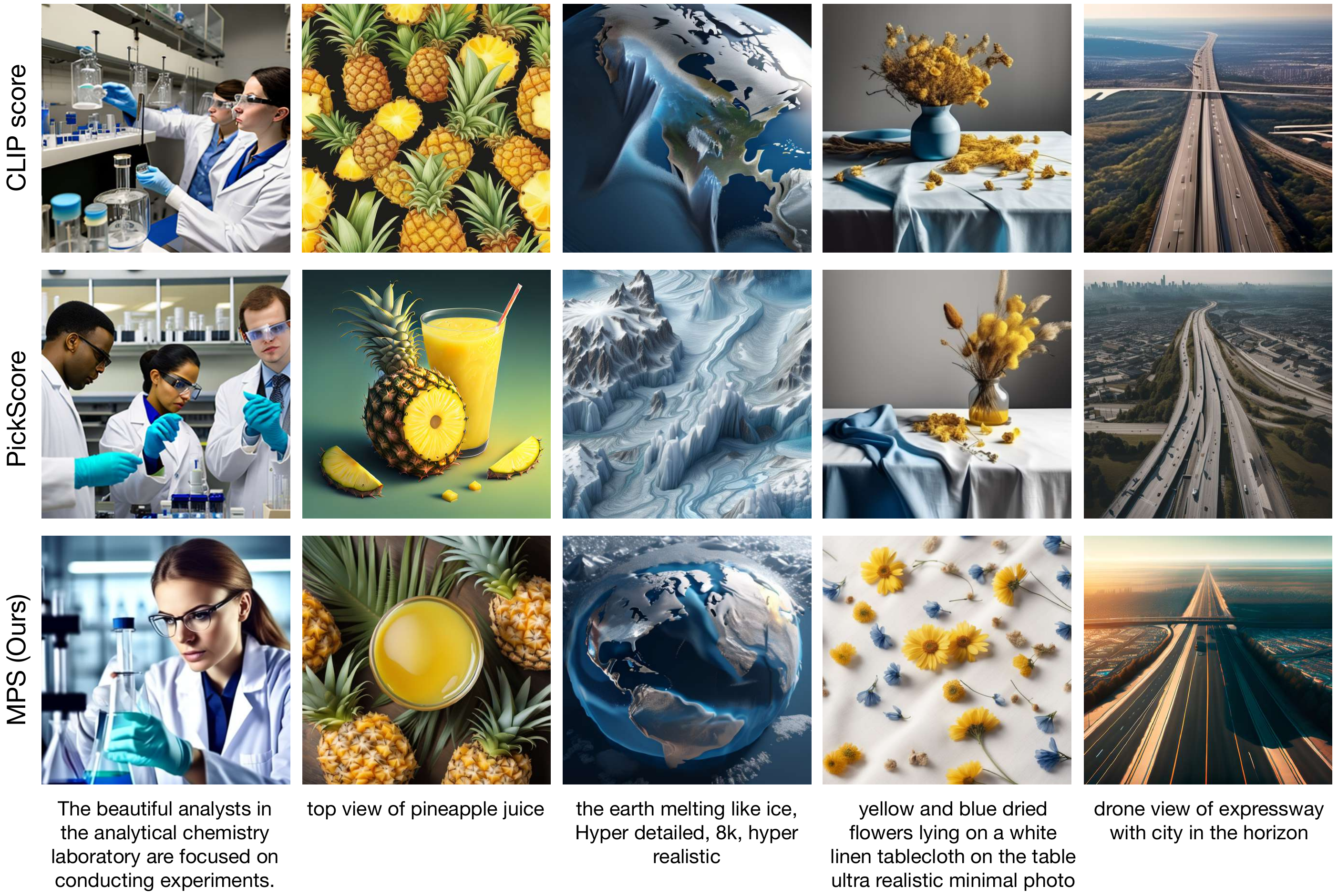}
\par\end{centering}
\vspace{-4pt}
\caption{Comparing images selected by CLIP Score \cite{CLIP_s}, PicScore \cite{PickScore}, and our MPS.
}
\label{fig:supp-reranking}
\vspace{-10pt}
\end{figure}

The score functions of text-to-image models serve to rank generated images by selecting generated images with higher scores to enhance the quality of text-to-image models. 
We extract some prompts from the validation set and generate 50 images for each prompt using the text-to-image models detailed in Tab. 2 of the main text. We compared the statistical method CLIP score \cite{CLIP_s}, the PicScore \cite{PickScore} which learns human preferences, and our MPS method. 
Since PicScore only learns overall preferences for each image, we only employ our MPS conditioned on the overall score a fair comparison. 
We display the results of how these three models rank images generated from the same prompts. Fig. \ref{fig:supp-reranking} shows the selected images with the highest scores according to each model. These qualitative comparisons indicate that our MPS is capable of selecting more preferable images than the baseline methods.

\section{Prompt category}
We annotate the categories of the collected prompts based on the categories of Parti \cite{parti}. We merge some categories of Parti \cite{parti} and obtain 7 categories. These 7 categories, along with their corresponding original Parti categories, are as follows: Characters (People), Scenes (Indoor scenes, Outdoor scenes), Objects (Vehicles, World knowledge), Animals (Animals), Plants (Produce \& plants), Arts (Artifacts, Arts, Illustrations), and Food (Food \& beverage).

Additionally, as Fig. 2 in the main text shows, the initially collected prompts demonstrate a long-tail distribution in categories. 
Therefore, for categories with fewer prompts, we generate additional prompts leveraging Large Language Models (LLMs) to maintain a balanced distribution. Below are examples of generated prompts:

For the Animals category:
\begin{itemize}
  \item A peacock spreading its magnificent tail by the lakeside, attracting the attention of passing tourists who stop to admire.
  \item A group of polar bears frolicking in the snow, rolling and playing around.
  \item Two little rabbits chasing each other and playing on a grassy field, enjoying the warm sunshine.
  \item A group of ants transporting food to their nest in an organized line, showing hard work and diligence.
\end{itemize}

For the Plants category:
\begin{itemize}
  \item Sunlight filtering through the leaves, a bud ready to bloom sways in the gentle breeze.
  \item In the orchard, branches are laden with tempting sweet fruits.
  \item In autumn, the grapevines have turned yellow, with bunches of purple-red grapes glistening in the sunlight.
  \item A mountainside covered with red maple leaves, like a fiery beauty of autumn.
\end{itemize}

For the Arts category:
\begin{itemize}
  \item A little girl holding hands with a giant bear as they walk through an enchanted forest, with magical animals passing by occasionally.
  \item Original illustration in a cartoon style, cute and suitable for a children's storybook cover.
  \item An ink wash painting employing the splashed ink technique to depict the ambiance of mountains and rivers, conveying a majestic and grand atmosphere.
  \item An impressionist oil painting depicting the leisurely ambiance of an afternoon, with sunlight cascading over a beautiful garden.
\end{itemize}

For the Food category:
\begin{itemize}
  \item Delicious and crispy fried chicken.
  \item Green broccoli with a vegetable salad.
  \item Spicy and appetizing Korean kimchi.
  \item Sweet and sour, delicious strawberry milkshake.
\end{itemize}

\begin{figure*}[ht]
\begin{centering}
\includegraphics[width=0.96\textwidth]{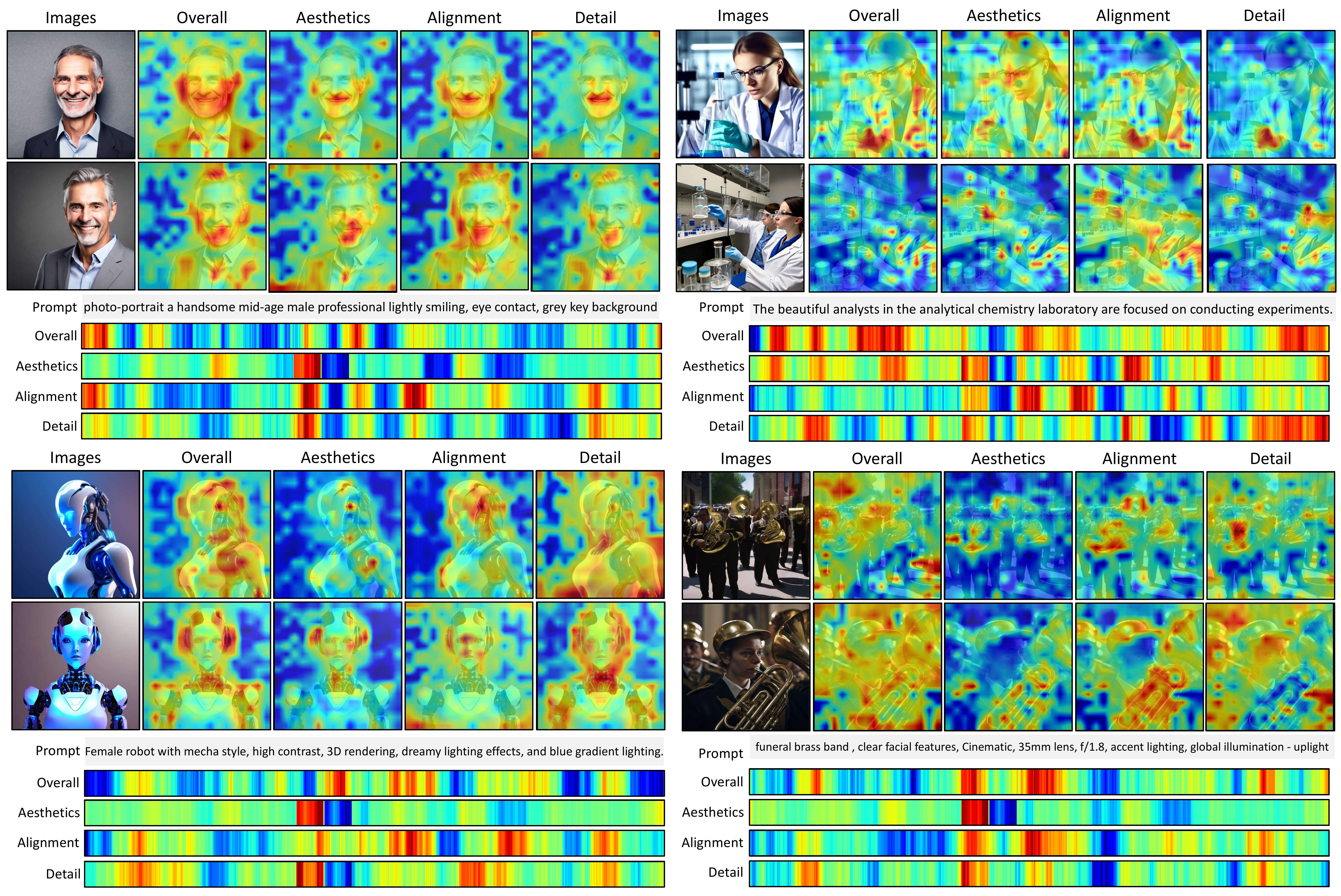}
\par\end{centering}
\vspace{-4pt}
\caption{More visualizations. Areas with a redder color indicate regions that MPS focuses on more when scoring.
}
\label{fig:supp-demos}
\end{figure*}

\section{More Visualizations}
Fig. \ref{fig:supp-demos} shows more visualization results. We use Grad-CAM to visualize the words in the prompts and the regions in the images that MPS focuses on when scoring the generated images. With the help of the condition mask, when the Aesthetic condition is given, MPS tends to focus on words like 'handsome', 'high construct', and 'beautiful'. When alignment is the condition, MPS tends to focus on attributes (such as 'photo', '3D rendering'), quantities ('35mm lens'), and locations ('chemistry laboratory'). Meanwhile, under the detail condition, MPS tends to focus on specific regions of the image (such as 'face', 'hair', 'hands' and 'limbs'). The visual results demonstrate that MPS can focus on different regions of the image and prompt according to different conditions, thereby predicting varying human preferences under different conditions.


{
    \small
    \bibliographystyle{ieeenat_fullname}
    \bibliography{main}
}
